\documentclass[final,5p,times,twocolumn]{elsarticle}

\usepackage{algorithm}
\usepackage{multirow}
\usepackage[noend]{algpseudocode}
\usepackage{amssymb}
\usepackage{graphicx}
\usepackage{amsmath}
\usepackage{subcaption,mwe}
\usepackage{soul}
\usepackage{xcolor}
\usepackage{amsfonts,bm}
\usepackage{textcomp}
\usepackage[hidelinks]{hyperref}
\graphicspath{ {./images/} }

\journal{Neurocomputing}

\begin{document}

\begin{frontmatter}



\title{ Evolutionary bagging for ensemble learning}


\author[inst1]{Giang Ngo}

\affiliation[inst1]{organization={Transitional Artificial Intelligence Research Group, School of Mathematics and Statistics, University of New South Wales},%Department and Organization
            city={Sydney}, 
            country={Australia}}

\author[inst2]{Rodney Beard}

\affiliation[inst2]{organization={Transitional Artificial Intelligence Research Group},%Department and Organization
            city={Sydney}, 
            country={Australia}}
 
\author[inst1,inst3,inst4]{Rohitash Chandra}

\affiliation[inst3]{organization={UNSW Data Science Hub, University of New South Wales},%Department and Organization
            city={Sydney}, 
            country={Australia}}

\affiliation[inst4]{organization={Data Analytics for Resources and Environments, Australian Research Council - Industrial Transformation Training Centre (ARC-ITTC)}, 
            country={Australia}}

\begin{abstract}

Ensemble learning has gained success in machine learning with major advantages over other learning methods. Bagging is a prominent ensemble learning method that creates subgroups of data, known as bags, that are trained by individual machine learning methods such as decision trees. Random forest is a prominent example of bagging with additional features in the learning process. Evolutionary algorithms have been prominent for optimisation problems and also been used for machine learning. Evolutionary algorithms are gradient-free methods that work with a population of candidate solutions that maintain diversity for creating new solutions. In conventional bagged ensemble learning, the bags are created once and the content, in terms of the training examples, are fixed over the learning process. In our paper, we propose evolutionary bagged ensemble learning, where we utilise evolutionary algorithms to evolve the content of the bags in order to iteratively enhance the ensemble by providing diversity in the bags. The results show that our evolutionary ensemble bagging method outperforms conventional ensemble methods (bagging and random forests) for several benchmark datasets under certain constraints. We find that evolutionary bagging can inherently sustain a diverse set of bags without reduction in performance accuracy.

\end{abstract}




\end{frontmatter}


\section{Introduction}
\label{sec:sample1}

Motivated by the "no free lunch" theorem \cite{585893}, ensemble learning methods use different methods to aggregate the predictions of multiple machine learning models. This often leads to a significantly improved performance when compared to individual models \cite{10.5555/2381019}. Bagging and boosting are currently the most popular ensemble methods, with random forest and \textit{AdaBoost} being the most prominent implementations, respectively. These methods cover a wide range of applications including face recognition \cite{1613003, KUMAR2020273}, anomaly detection \cite{SINGH2020188} and medicine \cite{med1,med2}. Bagging \cite{10.1023/A:1018054314350} is a prominent ensemble method for training individual learners on arbitrary sampled subsets of the original training dataset. \textcolor{black}{The aggregation of multiple learners leads to a lower variance of the model while its bias may remain the same given the bias-variance decomposition of error for machine learning models. Given multiple models of the same machine learning algorithm trained on different training data, the bias of the machine learning algorithm is the similarity between the models' average prediction and the ground truth, and its variance is the difference between the predictions \cite{Kohavi1996BiasPV}}. Random forest \cite{10.1023/A:1010933404324} is a prominent implementation of bagging that uses decision trees and introduces additional features to the sampling process. During each split, a random forest will consider only a random subset of the features as opposed to bagging, where all features are considered. Extra-tree \cite{geurts:hal-00341932} is an extension with random splits for decision trees to further reduce the computational cost of determining how to split the data. Random forest and extra-trees allow for fewer correlated decision trees than those learnt by bagging, a desirable property that allows different features to be represented instead of being dominated by strong predictors \cite{Ho2002ADC}.

A problem with bagging is that the bias of an individual machine learner is mostly the same as the bias of the combined model as better performance is the sole result of reduced variance \cite{10.1214/aos/1031689014}. \textcolor{black}{For example, if decision trees are not suitable for a particular data, it will likely underfit the data and produce high bias error. In that case, an ensemble of decision trees will have the same bias error as one single decision tree.} As a result, if the individual learners achieve high bias (model under-fitting), those errors are carried over to the aggregated predictions. It is clear that one can employ more sophisticated machine learning models such as artificial neural networks, but these do not improve over decision trees for problems where both computational power and data are limited. A natural direction to solve this problem is to optimize for bags that are more representative than the bootstrapped bags. 
\textcolor{black}{Some researchers have attempted to select an optimal size for each bag using bootstrapping and claimed that using bags with the same size as the whole training data is inefficient \cite{FRIEDMAN2007669,10.1214/aos/1031689014,MARTINEZMUNOZ2010143}.} However, these work with optimal size of bags without actually considering the data within each bag. On the other hand, it has also been argued that  we need to optimize the data within each bag and focus  on the specific problem of  imbalanced data with either over or under sampling for the data labels   \cite{https://doi.org/10.1002/sam.10061,BLASZCZYNSKI2015529,evobaggingimbalanced}.

\textcolor{black}{Evolutionary algorithms (EAs) are a family of optimisation algorithms motivated by the theory of  evolution \cite{back1996evolutionary}. EAs feature a population of individuals (candidate solutions) that cooperate and compete  using a measure of error (fitness).} Over time (evolution), the algorithm selects individuals from the population to create the next generation of individuals with selection, crossover and mutation operators \cite{7955308, Freitas2008}. Due to its flexibility, EAs are adopted widely for  machine learning models for various  tasks, such as multi-task learning \cite{evomultitask,CHANDRA201721}, decision tree induction \cite{5928432},  image segmentation \cite{evoimgseg}, Bayesian optimisation   \cite{app9}, and training deep learning models such as recurrent neural networks (neuroevolution) \cite{chandra2012adapting}. Since an ensemble of individual learners also resemble a population of candidate solutions, several attempts have been made to apply EAs to enhance ensemble learning methods such as  1.) evolving bagged training samples in an class imbalanced scenario \cite{evobaggingimbalanced, 6793456, ROSHAN2020103319}, 2.) assigning weights for individual learners \cite{Sylvester2005EvolutionaryE, 1716816, e22091020}, 3.) maintaining diversity between learners \cite{WANG201398, 10.1145/1276958.1277317}, and 4.) feature selection \cite{10.1007/978-3-540-30217-9_114}.

\textcolor{black}{In the past, attempts have been made to improve bagging, where the content of the bags in the ensemble has been a major focus of investigation \cite{evobaggingimbalanced, 6793456, ROSHAN2020103319}}. We note that the methods that did not alter the training samples in each bag had the advantage of optimizing over a smaller search space \cite{Sylvester2005EvolutionaryE, 1716816, WANG201398, 10.1145/1276958.1277317, 10.1007/978-3-540-30217-9_114}; e.g. the set of weights for individual learners or the subsets of features \cite{Sylvester2005EvolutionaryE, 1716816, e22091020}. However, these methods lacked the ability to optimize the set of training samples in each bag as these remained the same throughout the evolution process. This could be a major disadvantage if an informative representation of the data were required, such as when over-sampling of the majority class or under-sampling of the minority class were needed in an imbalanced scenario. \textcolor{black}{Furthermore, during model learning from these bags (individual learners),} it is important  to alter the content of the bags so they are not similar to each other, which otherwise would lead to highly correlated learners. García and Herrera \cite{6793456} proposed a set of algorithms for under-sampling for classification with imbalanced datasets using evolutionary algorithms, by maximizing accuracy and minimizing the amount of data used. Sun et al. \cite{evobaggingimbalanced} applied a similar framework, with a more complicated fitness function to maintain a balanced ratio between classes and a set of diverse bags. Roshan and Asadi \cite{ROSHAN2020103319} enforced these properties along with good classification performance using a multi-objective optimization process. \textcolor{black}{Although promising results were achieved, the ensemble featured considerable randomness, due to the crossover operator which executes} on a \textcolor{black}{random predefined probability distribution instead of a directed optimization that prioritises individual learners with higher classification performance.} 
The lack of directed optimization strategy cannot guarantee a stable learning process especially with massive datasets. In addition, these studies \cite{6793456, evobaggingimbalanced, ROSHAN2020103319} are limited to under-sampling in imbalanced classification problems, which is prone to biased selection of the majority class and also a potential loss of important data if care is not taken.
 
In our paper, we propose an  evolutionary bagged ensemble learning framework where we utilise evolutionary algorithms to \textcolor{black}{shuffle and update the data amongst} the bags in order to iteratively enhance the diversity of the ensemble. Our method optimizes a simple objective where crossover operator is designed so that we are able to modify the set of training samples in a bag given the performance of its corresponding learner. The other components of our method inherently maintains diversity between bags and attempts to decrease the errors associated with bias in each bag. We provide extensive evaluation on both balanced and imbalanced datasets for classification problems. Finally, we compare the performance of our method with other ensemble methods, such as the canonical bagging  and random forests. 

The remainder of the paper is organized as follows: Section 2 reviews bagging and evolutionary algorithms. Section 3 introduces our framework with a detailed explanation of its components. Section 4 presents results with benchmark datasets and Section 5 provides a discussion. Finally, Section 6  provides the conclusions.

\section{Background}

Genetic algorithms \cite{10.5555/230231} are prominent evolutionary algorithms that evolve a population of candidate solutions using evolutionary operators. \textcolor{black}{Initially, genetic algorithms used binary encoding \cite{10.5555/230231} and later real-coded genetic algorithms were developed \cite{Goldberg1991RealcodedGA}. Binary encoding suffers from  computational challenges, accuracy, and search space discontinuities \cite{CARUANA1988153}.   Real-coded encoding \cite{real1, real2, real3} have shown better performance when compared to binary  encoding for selected problems \cite{realga}.}
The overall framework of genetic algorithm is similar to other evolutionary algorithms such as evolution strategies \cite{evostrategies}, differential evolution \cite{diffevo}, and particle swarm optimisation \cite{488968}. The major differences is in the way a new population is created using different genetic operators. The fitness score (error) measures the quality of the solutions and is the key measure by which the evolutionary operators select individuals to create new solutions \cite{back1996evolutionary}. \textcolor{black}{A candidate solution includes a set of parameters that needs to be estimated for a problem (model). The goal of the optimization process is to search for a candidate solution with the highest fitness score.}

The ability to search for the global optimal solution using genetic algorithm has led to their wide application in numerous areas. Some of its popular applications include evolutionary game theory \cite{app1}, scheduling applications \cite{app2}, image processing \cite{app3}, traveling salesmen problem \cite{app4}, vehicle routing problem \cite{app5}, and many multi-modal optimization tasks \cite{app6, app7, app8}. In machine learning, the most common use of genetic algorithms is for hyperparameter tuning; where the search space is  small, and its applications range from Bayesian optimisation \cite{app9}, convolutional neural networks (CNNs) \cite{app12}, long short-term memory (LSTM) networks \cite{app10}, and fuzzy logic classification \cite{app11}.

\section{Methodology}

Our evolutionary bagged ensemble learning  method \textcolor{black}{(referred to as EvoBagging, hereforth) employs an evolutionary algorithm to improve bag samples generated from bagging. In bagging, bootstrapping iteratively samples} a dataset with replacement until a desirable size is reached and is commonly used to control the stability of the results in statistical learning \cite{efron1994introduction}. \textcolor{black}{The content of all bags are initialised at the beginning and remain unchanged.}
\textcolor{black}{EvoBagging evolves the content of all bags by applying the crossover and mutation operators prioritising bags (and their corresponding trained individual learners) with a higher classification performance.}   
We also introduce a generation gap with selection to help maintain diversity between \textcolor{black}{individual learners and to remove bags with poor performance}.  

%Moreover, we do not use the selection operator as a separate step to produce new offspring (selection is still used to select parents for crossover) in order to promote diversity with the crossover operator. The separate selection operator in this case would also hinder the bag evolution and shrinks data coverage \textcolor{black}{(i.e. the percentage of the total training data that is present in all bags)} when it is used with crossover.

\subsection{EvoBagging Algorithm}

\begin{figure*}
  \centering
  \includegraphics[scale=0.4]{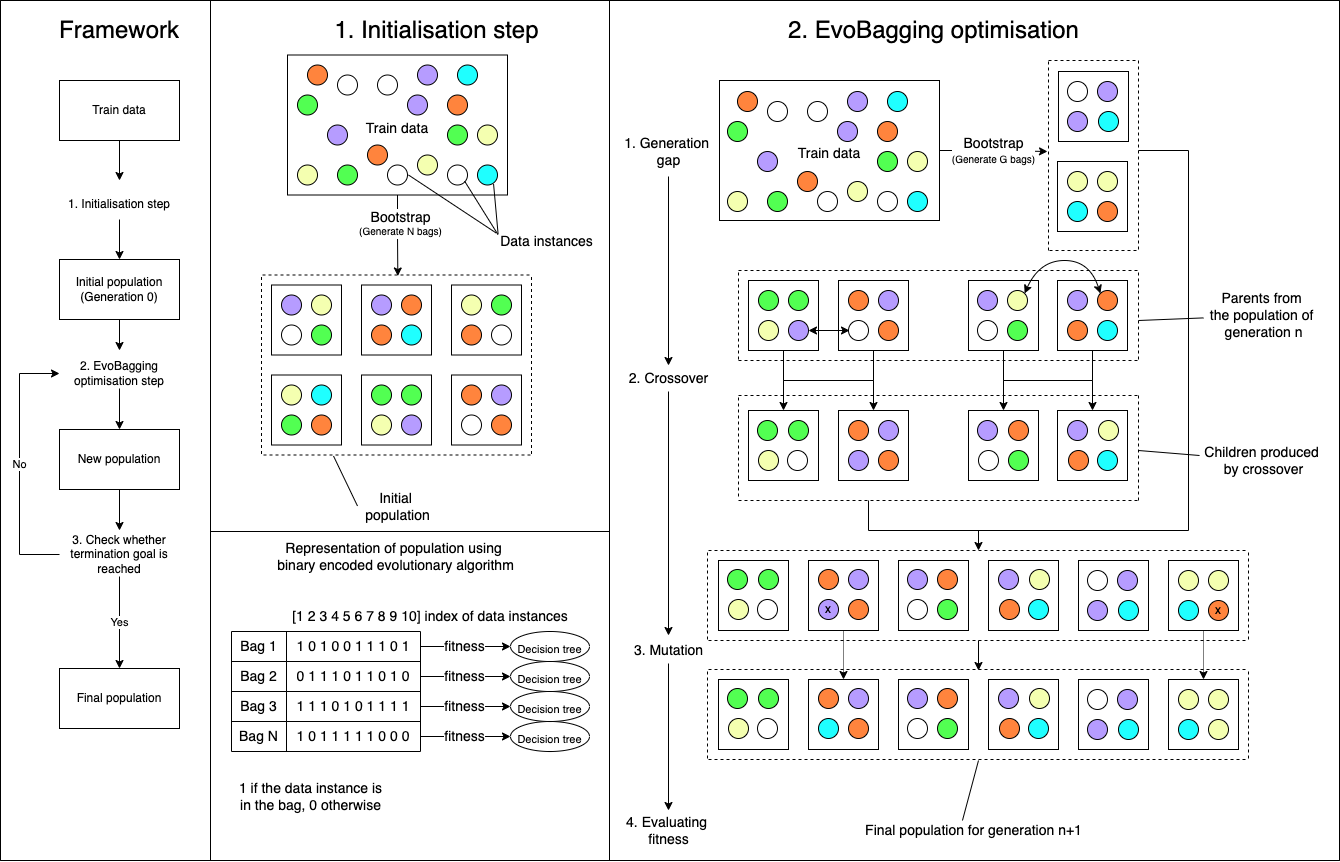}
  \caption{The EvoBagging framework where the bags are initialised with bootstrapping and iteratively optimised with EvoBagging. Note that the population size is determined by the number of bags in the ensemble  and the population representation (binary encoding) in the ensemble is explicitly shown. }
  \label{fig:EvoBagging}
\end{figure*}

The canonical bagged ensemble learning method (bagging) involves generating random subsets of training samples using bootstrapping and aggregating the results from individual models trained on those subsets. Bagging requires two hyper-parameters, i.e. the sample size $S$ and the number of bags $N$. Initially, bootstrapping is used to generate $N$ bags where each bag contains $S$ training samples. Each bag can be sampled uniformly from the original training set with replacement. A machine learning model can be subsequently trained with the bag's data and such an operation can also be implemented in parallel since the bagged models are independent. The final prediction from bagging is an aggregate of the predictions of all individual models, which is based on voting for classification and averaging for regression. In EvoBagging, we employ decision trees as the designated machine learning model for the bags, due to its training ability that takes low computational requirements when compared to other methods, such as artificial neural networks. Therefore, the individuals in the population of EvoBagging are bags that contain data instances. At every generation, the algorithm creates new population of individuals  using genetic operators such as crossover and mutation; hence, the data in the bags are changed over time. 

The major goal of evolutionary algorithms is to find the best solution for an optimization problem. \textcolor{black}{ In the case of EvoBagging (as shown in Figure 1 and Algorithm 1), the general goal remains the same, but its purpose is slightly different since the application is machine learning rather than optimisation. We are more interested in the overall performance of the population that is represented by the fitness given by the bags (individual learners), and the population diversity over time (evolution).} During evolution of the population, it is expected that the solutions within the population will also improve, and their errors will be reduced. Therefore, the fitness function used for optimization in the case of EvoBagging is still used, but the implicit goal is to improve the performance of the whole ensemble and take advantage of diversity of the individuals. In other words, the goal of EvoBagging is to   gradually improve the performance of all the individuals in the population which will eventually lead to a better performance of the ensemble. \textcolor{black}. We note that the individuals in the population represent the indices of the data in the bag and over time, the indices are evolved; hence, this is discrete parameter optimisation for individual learning models that learn from the data in the bags.

\begin{algorithm}[htbp!]
  \caption{EvoBagging}
  \textbf{Input}
  
  $term\_goal$: termination goal of the evolution process
  
  $N$: number of bags
  
  $S$: maximum bag size
  
  $G$: number of generation gap
  
  $M$: number of bags for mutation
  
  $MS$: mutation size
  
  \textbf{Output}
  
  Optimized solution for the problem
  \begin{algorithmic}[1]
    \label{alg:evobagging}
        \State $C=N-G$
        \State Generate initial population $P$ of $N$ bags of random size $s \in U(S/2, S)$
        \State Fit each bag and evaluate fitness
        \While{$term\_goal$ not reached} 
        \State Start new blank population $P'$
        \State Add $G$ new random bags to $P'$ as generation gap
        \For{i in $1:C/2$}
            \State Choose 2 parent bags from the population
            \State Perform crossover on these two bags
            \State Add 2 new children to $P'$
        \EndFor
        \State Mutate $M$ random bags in $P'$ with mutation size $MS$
        \State Update $P \leftarrow P'$
        \State Fit each bag again and evaluate fitness
        \EndWhile
  \end{algorithmic}
\end{algorithm}

The algorithm begins by generating $N$ bags as the initial population $P$, \textcolor{black}{i.e the number of individuals in $P$ will be same as number of bags.} We focus on improving the data samples by creating diversity in each bag, and the number of bags ($N$) will not change with evolution. \textcolor{black}{However, the number of data samples (items) in the respective bags are determined heuristically.} Note that we use a binary encoded evolutionary algorithm where each gene in the chromosome (individual) represents whether the data instance is in the bag as shown in Fig 1. Note that the number of genes in the individual is set to the maximum size of the bags which is dependent on the size of training data.  The evolutionary algorithm at every generation generates the bags using bootstrap with replacement and then uses evolutionary operators to enhance the quality of the bags. In the first generation, the bag size is determined arbitrarily  $s \in U(S/2, S)$, where $S$ determines the maximum size of a bag. The samples in each bag are selected using simple random sampling with replacement from the training set. We then train the decision tree model using the data from each bag and assign the resulting fitness to the bag. Each new generation begins with $G$ new bags generated randomly using bootstrapping to create a generation gap in the population $P'$. The algorithm then proceeds to produce offspring by performing crossover on $N - G$ bags selected from the last generation (so that the new bags from generation gap are not involved in this selection). These new individuals ($P'$) then become the current population $P=P'$.  We implement the mutation operator  on $M$ arbitrary selected bags from the updated population $P'$. The generation completes by fitting each bag using the individual learner (decision tree) and then assigns the fitness score to all the individuals in $P'$. \textcolor{black}{The algorithm stops if a termination condition is reached such as a user-defined fitness threshold, number of iterations, or classification performance of the ensemble.}

\subsection{Fitness function}

After fitting a bag with a pre-determined machine learning model, the fitness of the bag will depend on the bag  size ($\phi$)  as given below:

\begin{equation}
    fitness(bag) = \alpha \times \frac{K +  \phi_b}{K}
\end{equation}

\textcolor{black}{where, $\alpha$ is the classification performance of the model associated with the bag ($b$) in the ensemble, and $K$ is a user-defined hyper-parameter} for encouraging larger bags. The prediction of each training sample is used later, i.e. during the crossover operation. Apart from the classification performance, we add a simple term to create a bias for the bags with more samples. This is because optimizing only the classification accuracy  metric will likely lead to small over-fitted bags. The additional bags in the ensemble can be seen as a remedy for this issue, but it would also require complicated procedures to control the diversity of the bags.

\subsection{Generation gap}

\textcolor{black}{We introduce new bags in the ensemble at every generation}, where each bag is generated similar to the initial bags with a random size $s \in Uniform(S/2, S)$. The purpose of the generation gap is to replace bags with weaker performance and to enforce a diverse population. While the latter is an immediate consequence of the generation gap, the former can be seen as a result of both generation the gap and crossover. The reason is that crossover will probabilistically select the parent bags based on their fitness, which means bags with higher fitness are more likely to be selected. The remaining bags of the next generation are produced by the generation gap which replaces bags with lower fitness with new ones.

\subsection{Crossover and mutation}

The goal of the crossover operator is to ``diversify" the content of the bag so that the model fitted on its data has better performance. We simplify the step of selecting $C$ crossover parents by using a rank selection scheme where the offspring with the highest fitness are selected for crossover (after the generation gap scheme). The number of offspring (new solutions) is equal to the number of parents used for crossover. Given $C$ selected crossover parents, each pair of new offspring is created by recombining two random parents. In particular, the parents will be randomly allocated to $C/2$ pairs where each pair will exchange their training samples. If the prediction for each training sample in a bag in the evaluation step is accurate, it will remain in that bag. Otherwise, it is transferred to the other parent bag.

As for mutation, $M$ bags will be selected randomly from the current population $P'$. Let $B$ be one of the selected bag, and $B^c$ be the set of all training samples that are not in $B$. Bag $B$ will be mutated by replacing selected random samples given by the mutation size ($MS$) from $B$ with the same amount of random samples from $B^c$. This operator aims to maintain the diversity of bags in the population without complicated computation.

\section{Results}

This section evaluates the EvoBagging algorithm on several benchmark machine learning  datasets. The evaluation focuses on classification problems for both class balanced and imbalanced datasets.

\subsection{Datasets}

\subsubsection{N-bit parity and two-spiral problem}

%The algorithm proposed is first tested on n-bit parity and two-spiral classification problems. 

The n-bit parity problem has been used as a standard problem for machine learning algorithm as demonstrated in earlier works \cite{nbitrohitash,572107}. The objective of the n-bit parity problem is simply to learn a mapping function which determines whether the sum of a binary vector is odd or even. We present the 3-bit parity problem in Table \ref{tab:nbitexample}, where we show the input binary string (A, B, and C) and corresponding odd parity bit.

\begin{table}[htbp!]
\small
    \caption{3-bit parity dataset}
    \centering
    \begin{tabular}{|c|c|c|c|}
         \hline
         \hline
         A & B & C & Odd parity bit \\
         \hline
         \hline
         0 & 0 & 0 & 1 \\
         0 & 0 & 1 & 0 \\
         0 & 1 & 0 & 0 \\
         1 & 0 & 0 & 0 \\
         0 & 1 & 1 & 1 \\
         1 & 0 & 1 & 1 \\
         1 & 1 & 0 & 1 \\
         1 & 1 & 1 & 0 \\
         \hline
         \hline
    \end{tabular}
    \label{tab:nbitexample}
\end{table}

\begin{figure}[h]
\centering
\includegraphics[scale=0.6]{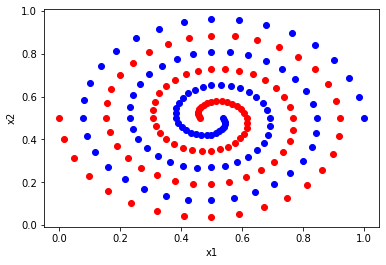}
\caption{Visualization of two-spiral data with tangled partitions. Note that blue and red dots represent the two classes for data instances given by 2 features (x1 and x2).}
\label{fig:two-spiral}
\end{figure}

Similarly, we conduct an experiments using the 6-bit and 8-bit parity problems, which have 64 and 256 samples (binary strings) respectively. Note that in these problems, we report only the training accuracy, since there is no test dataset available. The two-spiral problem is considered  to be a relatively  challenging problem for binary classification. We generate 194 samples (with two features, x1 and x2) for this problem and provide a  visualization of the data  in Figure \ref{fig:two-spiral}.

\subsubsection{Benchmark datasets}

We utilize eight commonly used classification datasets from the University of California - Irvine (UCI) machine learning repository \cite{UCIDatasets}. \textcolor{black}{In addition, we evaluate the performance of  EvoBagging on three real-world datasets provided by Penn Machine Learning Benchmarks} \cite{Olson2017PMLB}. In general, the each of the eleven datasets vary in size, number of features, and number of classes and details are shown in Table \ref{tab:uci}.

\begin{table}[htbp!]
\caption{\label{tab:table-name} Dataset information with class distribution (showing the proportion of the minority  and the majority classes).}
\centering
\small
 \begin{tabular}{|c|c|c|c|c|c|}
 \hline
 \hline
 Dataset & \#Rows & \#Features & \#Classes \\ 
 \hline
 \hline
 Red wine \cite{redwine} & 1599 & 11 & 2 (13.57\%, 86.43\%) \\ 
 Abalone \cite{UCIDatasets} & 4177 & 8 & 4 (6.25\%, 45.17\%) \\
 Breast cancer \cite{UCIDatasets} & 569 & 30 & 2 (37.26\%, 62.74\%) \\
 Pima \cite{UCIDatasets} & 768 & 8 & 2 (36.2\%, 63.8\%) \\
 Mnist \cite{UCIDatasets} & 1797 & 64 & 10 with 10\% each \\ 
 Car \cite{UCIDatasets} & 1728 & 6 & 4 (3.76\%, 70.02\%) \\
 Tic-tac-toe \cite{UCIDatasets} & 958 & 9 & 2 (34.66\%, 65.34\%) \\
 Ionosphere \cite{UCIDatasets} & 351 & 34 & 2 (35.9\%, 64.1\%) \\
 Churn \cite{Olson2017PMLB} & 5000 & 20 & 2 (14.14\%, 85.86\%)\\
 Ring \cite{Olson2017PMLB} & 7400 & 20 & 2 (49.51\%, 50.49\%)\\
 Flare \cite{Olson2017PMLB} & 1066 & 10 & 2 (17.02\%, 82.98\%)\\[1ex] 
 \hline
 \hline
 \end{tabular}
\label{tab:uci}
\end{table}

%xxxxxxxxxxxxxxxxxx Giang -- update from here

\subsection{Experiment setting}

\textcolor{black}{As noted earlier, we use decision trees for individual learners in the EvoBagging ensemble and provide further comparisons with mainly tree-based ensemble methods. We consider bagging, random forests, and extra-trees are the baseline models for comparison. The gradient boosting model is also included for a comparison of test classification performance on benchmark datasets.} 

In EvoBagging, the maximum training sample size for a new bag $S$ is the same as the size of the training data. To facilitate fair comparisons with other ensemble models, the number of bags in EvoBagging and all baselines is equal to the optimal number of bags for bagging. To find the optimal number of bags for bagging, we run a search with an interval of 10 and select the one with the highest test classification metric. As for stopping criteria, we terminate the EvoBagging training after a certain number of iterations which have been determined from trial experiments. We determine other hyper-parameters ($G$, $M$, $MS$, and $K$) by evaluating accuracy (5-fold cross-validation) on the training set as follows:

\begin{itemize}
  \item $G\in$ \{10\%, 15\%, 20\%, 25\%, 30\%\} of $N$
  \item $M\in$ \{5\%, 6\%, 7\%, 8\%, 9\%, 10\%\} of $N$
  \item $MS\in$ \{5\%, 10\%\} of $S$
  \item $K\in$ \{1000, 2000, 3000, ..., 20000\}
\end{itemize}

$MS$ is fixed at 5\% due to a better performance on all datasets. Further details on the influence of hyper-parameters are given in the ablation study (Section 4.4). The test set accounts for 20\% of the total original data and is randomised using stratified split in order to keep the same class distribution between the train set and the test set. We apply majority voting for EvoBagging and all baselines  
to aggregate the predictions of individual learners. We run 30 independent experiments with different random initialisation (content of bags) and report the mean and standard deviation of the respective metrics. Table \ref{tab:config} presents the details about the configuration for the benchmark datasets and two-spiral problem.

\begin{table}
\small
    \caption{Experimental configurations}
    \centering
    \begin{tabular}{|c|c|c|c|c|c|}
         \hline
         \hline
         Dataset & $N$ & $G$ & $M$ & $K$ & $iter$ \\
         \hline
         \hline
         Red wine & 50 & 20\% & 10\% & 5000 & 25 \\
         Abalone & 50 & 20\% & 10\% & 10000 & 35 \\
         Breast cancer & 20 & 20\% & 10\% & 2000 & 20 \\
         Pima & 60 & 15\% & 8\% & 2000 & 15 \\
         Mnist & 60 & 10\% & 5\% & 5000 & 20 \\
         Car & 60 & 15\% & 8\% & 5000 & 30 \\
         Tic-tac-toe & 70 & 20\% & 10\% & 4000 & 20 \\
         Ionosphere & 40 & 25\% & 10\% & 1000 & 20 \\
         Churn & 70 & 25\% & 6\% & 15000 & 35 \\
         Flare & 80 & 15\% & 6\% & 3000 & 40 \\
         Ring & 50 & 15\% & 10\% & 18000 & 15 \\
         Two-spiral & 40 & 25\% & 10\% & 1000 & 40 \\
         \hline
         \hline
    \end{tabular}
    \label{tab:config}
\end{table}

The n-bit parity problem is trained multiple times with a range of number of bags (from 10 to 100 bags for both problems). For simplicity, $G$ is  fixed at 20\% of $N$, and mutation is performed on 10\% of $N$ with mutation size 1. We define hyper-parameter for controlling bag size $K=100$ for two-spiral, and $K=500$ for 6-bit and 8-bit parity problems, respectively.

\subsection{Preliminary results}

\begin{figure}[!h]
\centering
\begin{subfigure}[t]{.48\textwidth}
    \centering
    \includegraphics[width=\textwidth]{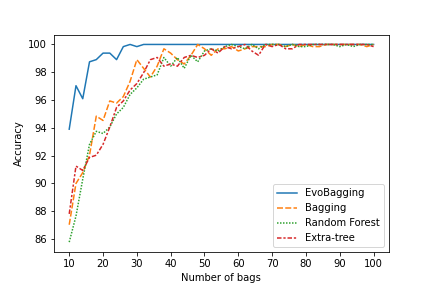}
    \caption{6-bit}
    \label{fig:6bit} 
\end{subfigure}
\begin{subfigure}[t]{.48\textwidth}
   \centering
   \includegraphics[width=\textwidth]{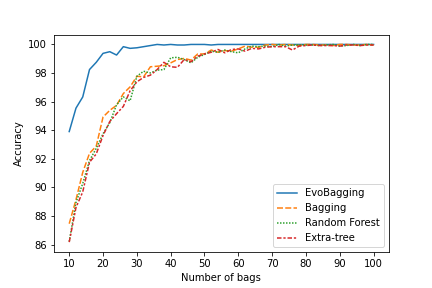}
   \caption{8-bit}
   \label{fig:8bit}
\end{subfigure}
\caption{Classification accuracy on the n-bit parity problems with varying number of bags}
\label{fig:nbitperf}
\end{figure}

\textcolor{black}{Figures \ref{fig:6bit} and \ref{fig:8bit} presents the training accuracy over number of bags for EvoBagging and the bagging-based baselines (i.e. bagging, random forests, and extra-trees). EvoBagging, initially shows better performance when compared to  bagging and random forest} for both 6-bit and 8-bit parity problems. The difference in training accuracy is more substantial with smaller number of bags. Moreover, all the methods eventually reach 100\% of training accuracy when more bags are added. This indicates that when the number of learners are constrained, EvoBagging can provide a considerably better result than other bagging-based ensemble methods.

\begin{table}[htbp!]
    \caption{Results for benchmark datasets and the two-spiral problem}
    \centering
    \small
    \begin{tabular}{|c|c|c|c|}
         \hline
         \hline
         Dataset & Model & Train & Test \\
         \hline
         \hline
         \multirow{5}{*}{Red wine} & Bagging & 99.95 (0.001) & 91.56 (0.005) \\
         & Random forest & 99.97 (0.001) & 92.0 (0.004) \\
         & ExtraTrees & 99.99 (0.001) & 92.5 (0.003) \\
         & Gradient Boosting & 94.37 (0.003) & 89.38 (0.005) \\
         & EvoBagging & 99.91 (0.001 & \textbf{92.76 (0.013)} \\
         \hline
         \multirow{5}{*}{Abalone} & Bagging & 99.96 (0.001) & 61.44 (0.004) \\
         & Random forest & 99.98 (0.002) & 62.34 (0.008) \\
         & ExtraTrees & 99.99 (0.001) & 60.65 (0.004) \\
         & Gradient Boosting & 71.45 (0.002) & 62.56 (0.007) \\
         & EvoBagging & 91.09 (0.002) & \textbf{63.97 (0.007)} \\
         \hline
         \multirow{3}{*}{Breast} & Bagging & 99.89 (0.001) & 96.06 (0.008) \\
         & Random forest & 99.97 (0.001) & 95.72 (0.009) \\
         \multirow{2}{*}{cancer} & ExtraTrees & 99.99 (0.001) & 97.26 (0.008) \\
         & Gradient Boosting & 99.56 (0.002) & 97.31 (0.010) \\
         & EvoBagging & 99.87 (0.002) & \textbf{97.75 (0.011)} \\
         \hline
         \multirow{5}{*}{Pima} & Bagging & 99.96 (0.002) & 75.22 (0.019) \\
         & Random forest & 99.99 (0.001) & 74.34 (0.013) \\
         & ExtraTrees & 99.99 (0.001) & 73.38 (0.013) \\
         & Gradient Boosting & 88.76 (0.002) & 74.03 (0.012) \\
         & EvoBagging & 99.92 (0.001) & \textbf{77.52 (0.010)} \\
         \hline
         \multirow{5}{*}{Mnist} & Bagging & 99.99 (0.001) & 95.13 (0.005) \\
         & Random forest & 99.99 (0.001) & 97.08 (0.004) \\
         & ExtraTrees & 99.99 (0.001) & \textbf{97.78 (0.006)} \\
         & Gradient Boosting & 99.99 (0.001) & 95.28 (0.008) \\
         & EvoBagging & 99.99 (0.001) & 96.36 (0.007) \\
         \hline
         \multirow{5}{*}{Car} & Bagging & 99.99 (0.001) & 97.87 (0.003) \\
         & Random forest & 99.99 (0.001) & 96.87 (0.009) \\
         & ExtraTrees & 99.99 (0.001) & 96.82 (0.010) \\
         & Gradient Boosting & 97.90 (0.003) & 97.40 (0.008) \\
         & EvoBagging & 99.99 (0.001) & \textbf{98.51 (0.006)} \\
         \hline
         \multirow{5}{*}{Tic-tac-toe} & Bagging & 99.99 (0.001) & 99.5 (0.006) \\
         & Random forest & 99.99 (0.001) & 96.87 (0.009) \\
         & ExtraTrees & 99.99 (0.001) & 98.96 (0.003) \\
         & Gradient Boosting & 97.78 (0.004) & 97.92 (0.008) \\
         & EvoBagging & 99.98 (0.001) & \textbf{99.99 (0.001)}  \\
         \hline
         \multirow{5}{*}{Ionosphere} & Bagging & 99.98 (0.001) & 94.17 (0.013) \\
         & Random forest & 99.99 (0.001) & 93.54 (0.007) \\
         & ExtraTrees & 99.99 (0.001) & 94.37 (0.015) \\
         & Gradient Boosting & 99.64 (0.002) & \textbf{95.77 (0.010)} \\
         & EvoBagging & 99.93 (0.002) & \textbf{95.77 (0.017)}  \\
         \hline
         \multirow{5}{*}{Churn} & Bagging & 99.98 (0.002) & 95.3 (0.006) \\
         & Random forest & 99.99 (0.001) & 95.50 (0.012) \\
         & ExtraTrees & 99.99 (0.001) & 93.20 (0.010) \\
         & Gradient Boosting & 96.65 (0.006) & 95.00 (0.011) \\
         & EvoBagging & 99.99 (0.001) & \textbf{96.32 (0.015)} \\
         \hline
         \multirow{5}{*}{Flare} & Bagging & 88.73 (0.007) & 80.37 (0.021) \\
         & Random forest & 88.73 (0.008) & 81.31 (0.019) \\
         & ExtraTrees & 88.73 (0.007) & 78.97 (0.023) \\
         & Gradient Boosting & 86.62 (0.005) & \textbf{82.24 (0.016)} \\
         & EvoBagging & 88.58 (0.008) & 81.69 (0.025) \\
         \hline
         \multirow{5}{*}{Ring} & Bagging & 99.99 (0.001) & 96.28 (0.003) \\
         & Random forest & 99.97 (0.003) & 94.05 (0.009) \\
         & ExtraTrees & 99.99 (0.001) & 96.89 (0.010) \\
         & Gradient Boosting & 95.00 (0.003) & 94.73 (0.010) \\
         & EvoBagging & 99.99 (0.001) & \textbf{98.20 (0.024)} \\
         \hline
         \multirow{5}{*}{Two-spiral} & Bagging & 99.92 (0.002) & 63.41 (0.035) \\
         & Random forest & 99.97 (0.001) & 59.82 (0.042) \\
         & ExtraTrees & 99.99 (0.001) & 64.10 (0.023) \\
         & Gradient Boosting & 92.26 (0.007) & 64.10 (0.021) \\
         & EvoBagging & 99.94 (0.002) & \textbf{71.79 (0.043)} \\
         \hline
         \hline
    \end{tabular}
    \label{tab:result}
\end{table}

\begin{figure}[!h]
\centering
\begin{subfigure}[t]{.23\textwidth}
    \centering
    \includegraphics[width=\textwidth]{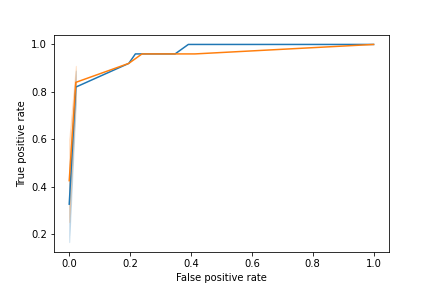}
    \caption{Class 0}
    \label{fig:class0} 
\end{subfigure}
\begin{subfigure}[t]{.23\textwidth}
   \centering
   \includegraphics[width=\textwidth]{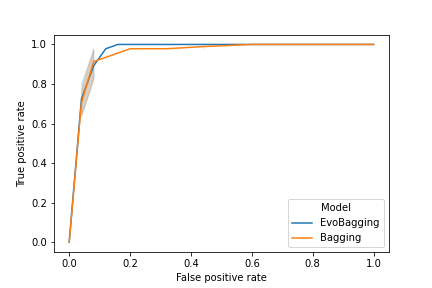}
   \caption{Class 1}
   \label{fig:class1}
\end{subfigure}
\caption{ROC curves for the Ionosphere dataset using Bagging and EvoBagging.}
\label{fig:roc_ion}
\end{figure}

\begin{figure}[htpb!]
\centering
\begin{subfigure}[t]{.23\textwidth}
    \centering
    \includegraphics[width=\textwidth]{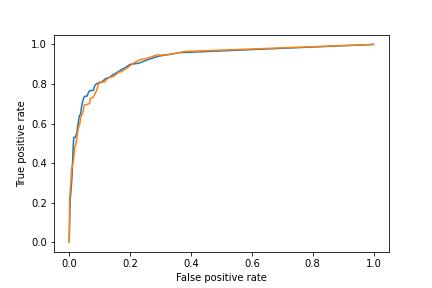}
    \caption{Class 1}
    \label{fig:class1} 
\end{subfigure}
\begin{subfigure}[t]{.23\textwidth}
   \centering
   \includegraphics[width=\textwidth]{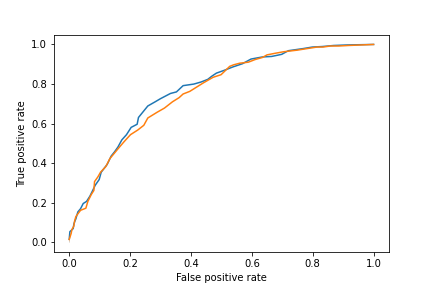}
   \caption{Class 2}
   \label{fig:class2}
\end{subfigure}
\begin{subfigure}[t]{.23\textwidth}
   \centering
   \includegraphics[width=\textwidth]{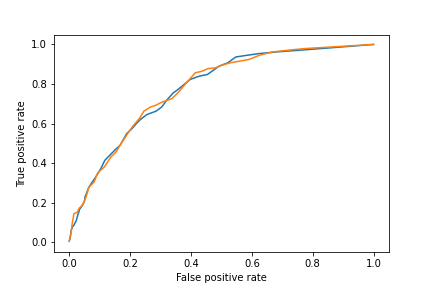}
   \caption{Class 3}
   \label{fig:class3}
\end{subfigure}
\begin{subfigure}[t]{.23\textwidth}
   \centering
   \includegraphics[width=\textwidth]{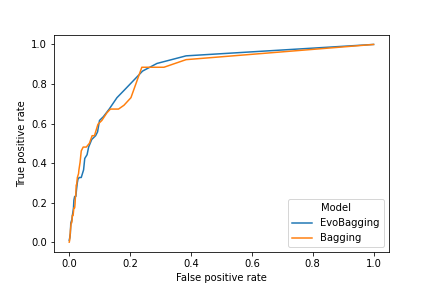}
   \caption{Class 4}
   \label{fig:class4}
\end{subfigure}
\caption{ROC curves for the Abalone dataset using Bagging and EvoBagging}
\label{fig:roc_aba}
\end{figure}

Table \ref{tab:result} shows classification accuracy of EvoBagging and all baselines for all benchmark datasets. We observe that EvoBagging successfully outperforms the baselines in most cases. In Figure \ref{fig:roc_ion} and \ref{fig:roc_aba}, we present the receiver operator characteristic (ROC) curves of bagging and EvoBagging, which shows comparable classification accuracy  for both methods. This result demonstrates its ability to improve the performance of methods related to bagging by building more representative sets of training samples. The two-spiral problem is challenging with non-linearly separable data, where EvoBagging  outperforms both bagging and random forests by 11.6\% and 18.3\% for accuracy, respectively.

\begin{figure}[h]
\centering
\includegraphics[scale=0.55]{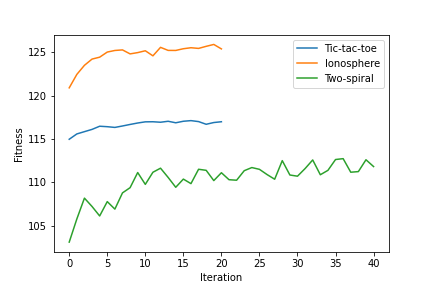}
\caption{Average fitness score of all bags for each iteration on three datasets. Note that the Tic-tax-toe and Ionesphere datasets have lower number of maximum iterations (20) when compared to the Two-spiral problem. }
\label{fig:fitness}
\end{figure}

Since the major feature of EvoBagging is to simultaneously improve the fitness scores of all bags, we record the average fitness score in Figure \ref{fig:fitness}, for some datasets instead of recording the maximum fitness score for each generation in traditional evolutionary algorithms. The bags gradually improve over the evolution process, and convergence is guaranteed after a few iterations (depending on the dataset).

\subsection{Influence of hyperparameters}

In this section, we experiment the sensitivity of EvoBagging on changes in the hyperparameters including maximum bag size $S$, generation gap $G$, number of mutated bags $M$, mutation size $MS$, and the bag size control $K$. We run these experiments are run on the Pima dataset with its best configuration specified in Section 4.1. Note that in the evaluation of a selected hyperparameter, all the other hyperparameters remain fixed during the experimental run.

\subsubsection{Maximum bag size $S$}

Table \ref{tab:max_size} presents the accuracy of EvoBagging, the average size of each bag, and the average depth of each decision tree (i.e. each individual learner) in the last iteration for four different ratios of the maximum bag size $S$ (i.e. 25\%, 50\%, 75\%., and 100\% of the number of samples in the training data).

\begin{table}[h]
    \caption{Comparing the ensemble's classification accuracy, average size of all bags in the final iteration, and average depth of individual decision trees between different values of $S$ for EvoBagging on Pima dataset}
    \centering
    \small
    \begin{tabular}{|c|c|c|c|}
         \hline
         \hline
         Ratio & Accuracy & Average bag size & Average tree depth \\
         \hline
         \hline
         25\% & 73.38 & 139.4 & 8.5 \\
         50\% & 75.97 & 284.7 & 11.1 \\
         75\% & 77.48 & 407.8 & 12.0 \\
         100\% & 77.52 & 547.1 & 12.4 \\
         \hline
         \hline
    \end{tabular}
    \label{tab:max_size}
\end{table}

As $S$ affects the initialisation of new bags in EvoBagging, a larger value of $S$ generally leads to a larger average bag size. It is observed that $S$ is proportional to the final average bag size in the last iteration. The size of each bag, in turn, affects the depth of the individual learner corresponding to that bag, so a larger $S$ also leads to a deeper decision tree. A ratio of 100\% generally produces the best classification result, but the gap is minimal compared to a ratio of 75\%.

\subsubsection{Generation gap $G$}

\begin{figure}[htbp!]
\centering
\includegraphics[scale=0.54]{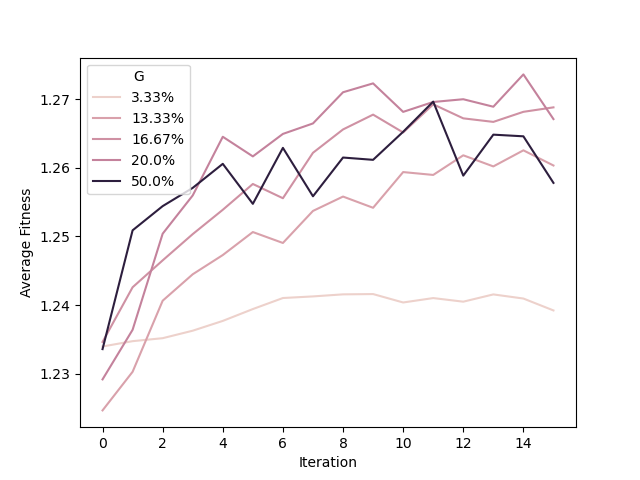}
\caption{The average fitness of all bags for each iteration of EvoBagging on Pima dataset with different $G$'s}
\label{fig:G}
\end{figure}

Figure \ref{fig:G} shows the average fitness of all bags for each iteration with different ratios of generation gap. The optimal value for $G$ produced by grid-search in this case is 16.67\% (for $N$=60). It can be seen that only minor changes happen when $G$ is 13.33\% or 20\% (i.e. close to the optimal value). When $G$ is 50\%, the average fitness fluctuates unpredictably due to replacing half of the population in each iteration. When $G$ is 3.33\%, the average fitness not only is lower but also improves quite slowly as a result of insufficient randomness added to the optimisation process, which is a main effect of the generation gap scheme.

\subsubsection{Number of mutated bags $M$}

\begin{figure}[htbp!]
\centering
\includegraphics[scale=0.54]{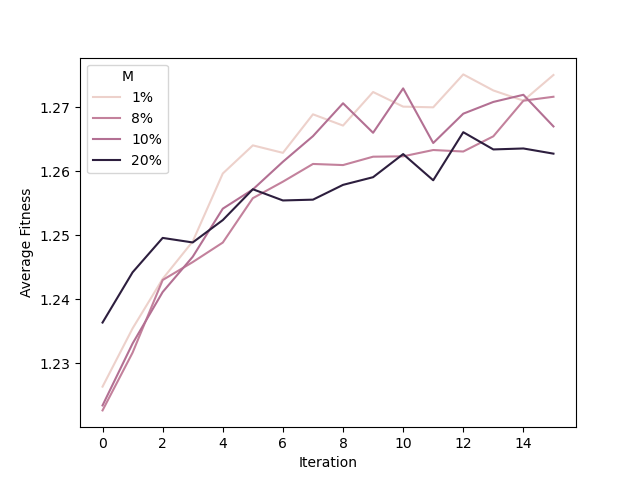}
\caption{The average fitness of all bags for each iteration of EvoBagging on Pima dataset with different $M$'s}
\label{fig:M}
\end{figure}

Figure \ref{fig:M} shows the average fitness of all bags for each iteration with different ratios for $M$. When the ratio is less than 10\%, the average fitness is mostly the same. However, a high ratio for $M$ like 20\% (i.e. one fifth of the bags are mutated in each iteration) leads to a lower average fitness as too much randomness is added to the population. Generally, the experiments have shown that the ratio for $M$ should be less than 10\% on all datasets.

\subsubsection{Mutation size}

\begin{figure}[htbp!]
\centering
\includegraphics[scale=0.54]{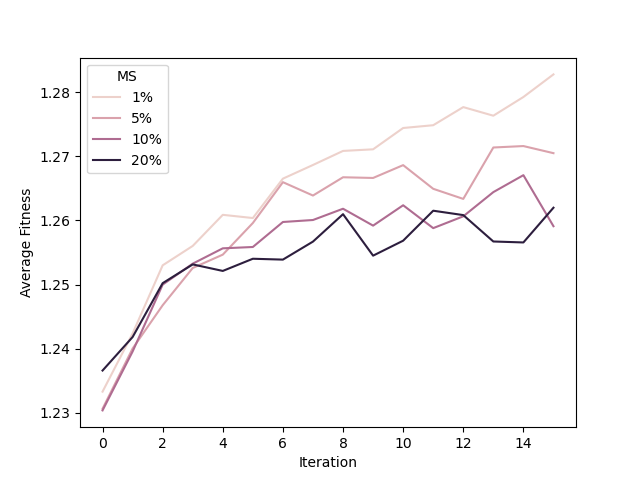}
\caption{The average fitness of all bags for each iteration of EvoBagging on Pima dataset with different $MS$'s}
\label{fig:MS}
\end{figure}

Figure \ref{fig:MS} shows the average fitness of all bags for each iteration with different ratio for mutation size given by $MS$. Generally, the average fitness evolves steadily when $MS$ is at 5\%. A larger ratio leads to lower fitness when a large proportion of the bag selected for mutation is altered. Note that when $MS$ is only 1\%, the average fitness is exceptionally higher when compared with other ratios. In that case, the bags selected for mutation remain mostly the same. However, such a small ratio is not favored as it limits the ability to induce diversity through randomly created content in the population.

\subsubsection{Bag size}

The purpose of the bag size given by $K$ is to give more score to larger bags so that the population does not collapse to multiple small bags that are easy to predict. Table \ref{tab:K} shows the classification accuracy and average bag sizes for different values of $K$. We observe that a smaller value of $K$ leads to larger bags, but larger bags do not necessarily correlate to better classification performance as over-represented training data instances may result in overfitted individual learners. The choice of $K$ should generally be proportionate to the size of the training data.

\begin{table}[h]
    \caption{Classification performance and bag sizes vary with different values of $K$}
    \centering
    \small
    \begin{tabular}{|c|c|c|}
         \hline
         \hline
         $k$ & Accuracy & Average bag size \\
         \hline
         \hline
         1000 & 74.03 & 550.7 \\
         2000 & 77.52 & 536.2 \\
         10000 & 75.97 & 531.9 \\
         20000 & 75.32 & 531 \\
         \hline
         \hline
    \end{tabular}
    \label{tab:K}
\end{table}

\subsection{Ablation study}

\subsubsection{Evolution}

A crucial objective of using an evolutionary algorithm in EvoBagging is to reduce biases of the individual learners while maintaining a comparable level of variance. We measure the bias in a machine learning model by the difference between prediction and the actual labels of the data. The bias can emerge from inappropriate learning algorithm and models, while the variance can emerge from sensitivity to noisy data by the machine learning model \cite{Kohavi1996BiasPV}. The bias-variance trade-off reflects on how well the model can generalise when given unseen (test) dataset. We consider the decomposition of mean square error (MSE) into bias squared and variance.

In this experiment, we measure bias for the task of binary classification given by:

\begin{equation}
    Bias = \begin{cases}
        0, & \text{if }  \hat{y}=y \\
        1, & \text{other wise.}
    \end{cases}
\end{equation}

\begin{figure}[htbp!]
\centering
\includegraphics[scale=0.55]{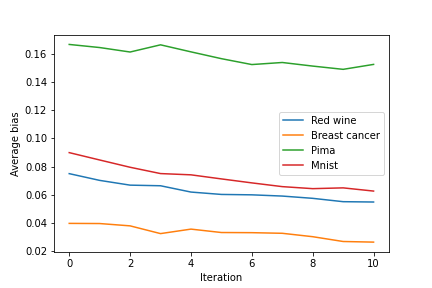}
\caption{The average bias of the individual learner (bag) in a population constantly reduces after each iteration on four different datasets.}
\label{fig:reduce-bias}
\end{figure}

In each bag of the ensemble, we obtain the average bias of the respective individual learner  for all the data samples in the test set. Figure \ref{fig:reduce-bias} shows the average of this measure for all individual learners in four different datasets. Overall, we observe that the bias gradually decreases over 10 generations of evolution. This provides important empirical evidence that EvoBagging is capable of reducing the bias error of individual learners (bags) with evolution.

Table \ref{tab:reduce-bias} shows the relative reduction of the average bias between the first and the last iterations on 12 benchmark datasets. Abalone, Pima, Flare and the two-spiral problem are the datasets with the lowest relative reduction. These are also the most difficult learning problem among these datasets considering the classification performances of all models shown in table \ref{tab:result}.

\begin{table}[h]
    \caption{The average biases of the first and last iterations of 12 benchmark datasets}
    \centering
    \small
    \begin{tabular}{|c|c|c|c|}
         \hline
         \hline
         Dataset & \multicolumn{2}{|c|}{Average bias} & \% reduced \\\cline{2-3}
         & First & Last & \\
         \hline
         \hline
         Red wine & 0.072 & 0.056 & 22.0\% \\
         Abalone & 0.326 & 0.318 & 2.5\% \\
         Breast cancer & 0.039 & 0.031 & 22.2\% \\
         Pima & 0.163 & 0.148 & 9.2\%\\
         Mnist & 0.091 & 0.061 & 32.2\%\\
         Car & 0.030 & 0.020 & 31.2\% \\
         Tic-tac-toe & 0.051 & 0.035 & 31.7\% \\
         Ionosphere & 0.061 & 0.041 & 32.0\% \\
         Churn & 0.042 & 0.033 & 20.7\% \\
         Flare & 0.152 & 0.147 & 3.3\% \\
         Ring & 0.073 & 0.063 & 13.8\% \\
         Two-spiral & 0.456 & 0.446 & 2.2\% \\
         \hline
         \hline
    \end{tabular}
    \label{tab:reduce-bias}
\end{table}

\subsubsection{The bias-variance trade-off}

It is of significant importance for EvoBagging to reduce the bias of individual learners (bags). However, it must not be achieved by sacrificing variance in order to eventually obtain a stable aggregated performance. In this experiment, we evaluate if EvoBagging can maintain a level of variance comparable to bagging by training on different training data. The purpose of the experiment is to compare the sensitivity of EvoBagging and bagging given changes in the training data. To be specific, we run each method 30 times, each time with different training data (content of bags) while keeping the same test data. The whole original dataset $D$ is split into a training set $D_{train}$ and a test set $D_{test}$ with a ratio, 80:20. The test set $D_{test}$ remains the same for all runs. In each run, a training set $D_{train}^i$ is formed by bootstrapping $D_{train}$ and has the same size as $D_{train}$. Table \ref{tab:datasetup} illustrates the setup of the data for this experiment.

\begin{table}
\small
    \caption{Illustration of data setup for measuring variance of the model}
    \centering
    \begin{tabular}{|c|c|}
         \hline
         \hline
         Instances in the original dataset $D$ & 1,2,3,4,5,6,7,8,9,10\\
         \hline
         \hline
         Instances in the test set $D_{test}$ & 9,10\\
         \hline
         \hline
         Instances in $D_{train}^1$ & 1,2,3,3,4,4,5,6\\
         Instances in $D_{train}^2$ & 2,2,3,4,5,6,7,8\\
         ... & ...\\
         Instances in $D_{train}^{30}$ & 1,2,2,3,3,5,7,7\\
         \hline
         \hline
    \end{tabular}
    \label{tab:datasetup}
\end{table}

\textcolor{black}{To measure the sensitivity of EvoBagging and bagging to different training data, we employ six diversity measures implemented by Albukhanajer et al. \cite{ALBUKHANAJER2017316}. It should be noted that these were originally used as measures for diversity between individual learners within an ensemble, which should be as diverse as possible. This means the predictions of the individual learners within an ensemble are expected to be different from each other. However, in this experiment, there are two differences. Firstly, the targets of the diversity measures are the ensembles trained by EvoBagging or bagging on different training data, not the individual learners within one single ensemble. Secondly, instead of a more diverse group as expected in the original implementation, our goal in this experiment is a less diverse group. Such a group of ensembles trained on different training data will make similar predictions which is equivalent to having a lower variance (i.e. a lower sensitivity to changes in the training data). In short, a more favourable result in this experiment is a less diverse model.}

Table \ref{tab:avgvar} presents the six diversity measures between 30 ensembles, each trained on a different training data $D_{train}^i$'s, for both EvoBagging and bagging. These measures are calculated on the predictions of each ensemble for the same $D_{test}$. %variance of the prediction on test data for  30 independent experimental  runs for five binary classification problems. 
We find that EvoBagging exhibits a similar level of variance \textcolor{black}{(diversity)} in comparison to bagging \textcolor{black}{(and even less varied on Red wine dataset)}. This investigation, along with the one about the ability to reduce bias shown in Section 4.5.1, explains why EvoBagging manages to reach a better performance than the counterparts. It is due to reducing biases for individual learners and maintaining low variance for the whole ensemble. \textcolor{black}{While maintaining low variance is an obvious result inherited from the aggregation step of bagging, reducing biases for individual learners is only achieved by optimizing a fitness function dedicated to this purpose.} %While the latter (EvoBagging) is an obvious result inherited from the aggregation step of bagging, the former is only achieved by optimizing a fitness function dedicated to this purpose.

\begin{table}[htpb!]
    \small
    \caption{Compare diversity measure among ensembles trained on different training data for both bagging and EvoBagging. In this experiment, being less diverse is desirable.}
    \centering
    \begin{tabular}{|c|c|c|c|}
         \hline
         \hline
         Dataset & Measure & Bagging & EvoBagging \\
         \hline
         \hline
         & Q statistics $\downarrow$ & 0.798 & 0.838 \\
         \multirow{4}{*}{Red} & Disagreement $\uparrow$ & 0.195 & 0.181 \\
         & Double fault $\downarrow$ & 0.112 & 0.115 \\
         \multirow{2}{*}{wine} & Kohavi-Wolpert variance $\uparrow$ & 0.088 & 0.081 \\
         & Entropy $\uparrow$ & 0.219 & 0.202 \\
         & Generalized diversity $\uparrow$ & 0.464 & 0.440 \\
         \hline
         \multirow{6}{*}{Ring} & Q statistics $\downarrow$ & 0.529 & 537 \\
         & Disagreement $\uparrow$ & 0.501 & 0.501 \\
         & Double fault $\downarrow$ & 0.207 & 0.207 \\
         & Kohavi-Wolpert variance $\uparrow$ & 0.226 & 0.226 \\
         & Entropy $\uparrow$ & 0.759 & 0.759 \\
         & Generalized diversity $\uparrow$ & 0.547 & 0.548 \\
         \hline
         \multirow{6}{*}{Mnist} & Q statistics $\downarrow$ & 0.591 & 0.599 \\
         & Disagreement $\uparrow$ & 0.323 & 0.327 \\
         & Double fault $\downarrow$ & 0.646 & 0.641 \\
         & Kohavi-Wolpert variance $\uparrow$ & 0.145 & 0.147 \\
         & Entropy $\uparrow$ & 0.383 & 0.388 \\
         & Generalized diversity $\uparrow$ & 0.200 & 0.203 \\
         \hline
         \multirow{6}{*}{Car} & Q statistics $\downarrow$ & 0.531 & 0.448 \\
         & Disagreement $\uparrow$ & 0.386 & 0.385 \\
         & Double fault $\downarrow$ & 0.217 & 0.216 \\
         & Kohavi-Wolpert variance $\uparrow$ & 0.174 & 0.173 \\
         & Entropy $\uparrow$ & 0.513 & 0.511 \\
         & Generalized diversity $\uparrow$ & 0.471 & 0.472 \\
         \hline
         \hline
    \end{tabular}
    \label{tab:avgvar}
    (More diverse if the measure is greater ($\uparrow$) or lower ($\downarrow$))
\end{table}

\subsubsection{Diversity between evolved bags}

In the design of EvoBagging, mutation and generation gap aims to keep the fitted individual learners diverse. That is, those learners must not be highly correlated and must not be highly similar to each other. It is essential to verify whether EvoBagging can be compared with other baselines regarding this property. %Since there are no direct methods to measure the similarity of decision trees (not to mention other complicated machine learning models), prediction variance between individual learners serves as an explicit indication of correlation.
Again, we use the six diversity measures in the previous section to measure the similarity between individual learners. Contrary to the previous section, in this experiment, a more favourable result is a more diverse group of individual learners.

\textcolor{black}{Table \ref{tab:diversity} shows the six diversity measures between individual learners for EvoBagging and bagging. We observe that EvoBagging is consistently more diverse on the three datasets Ring, Mnist and Car, while being comparable with bagging on Red wine.} %Table \ref{tab:diversity} shows the average prediction variance between individual learners for Bagging and EvoBagging. In the respective datasets for binary classification, we observe that the two methods are comparable. 
Therefore, the learners fitted using EvoBagging can be seen as relatively diverse, which demonstrates the effectiveness of the algorithm design. The mutation and generation gap successfully introduces further diversity to the ensemble of bags. We find that the ensemble usually shrinks to covering only a small part of the whole training dataset if further evolution with crossover operator is continuously applied. This is a result of breeding between only a group of bags with high fitness. 

\begin{table}[htpb!]
\small
    \caption{Compare diversity measures among individual learners between bagging and EvoBagging. In this experiment, being more diverse is desirable.}
    \centering
    \begin{tabular}{|c|c|c|c|}
         \hline
         \hline
         Dataset & Measure & Bagging & EvoBagging \\
         \hline
         \hline
         & Q statistics $\downarrow$ & 0.747 & 0.753 \\
         \multirow{4}{*}{Red} & Disagreement $\uparrow$ & 0.164 & 0.161 \\
         & Double fault $\downarrow$ & 0.062 & 0.0.061 \\
         \multirow{2}{*}{wine} & Kohavi-Wolpert variance $\uparrow$ & 0.080 & 0.079 \\
         & Entropy $\uparrow$ & 0.220 & 0.217 \\
         & Generalized diversity $\uparrow$ & 0.570 & 0.570 \\
         \hline
         \multirow{6}{*}{Ring} & Q statistics $\downarrow$ & 0.658 & 0.639 \\
         & Disagreement $\uparrow$ & 0.176 & 0.181 \\
         & Double fault $\downarrow$ & 0.049 & 0.049 \\
         & Kohavi-Wolpert variance $\uparrow$ & 0.087 & 0.089 \\
         & Entropy $\uparrow$ & 0.232 & 0.240 \\
         & Generalized diversity $\uparrow$ & 0.641 & 0.648 \\
         \hline
         \multirow{6}{*}{Mnist} & Q statistics $\downarrow$ & 0.734 & 0.726 \\
         & Disagreement $\uparrow$ & 0.184 & 0.189 \\
         & Double fault $\downarrow$ & 0.079 & 0.080 \\
         & Kohavi-Wolpert variance $\uparrow$ & 0.090 & 0.093 \\
         & Entropy $\uparrow$ & 0.256 & 0.260 \\
         & Generalized diversity $\uparrow$ & 0.537 & 0.543 \\
         \hline
         \multirow{6}{*}{Car} & Q statistics $\downarrow$ & 0.867 & 0.847 \\
         & Disagreement $\uparrow$ & 0.060 & 0.066 \\
         & Double fault $\downarrow$ & 0.016 & 0.017 \\
         & Kohavi-Wolpert variance $\uparrow$ & 0.029 & 0.033 \\
         & Entropy $\uparrow$ & 0.080 & 0.092 \\
         & Generalized diversity $\uparrow$ & 0.649 & 0.659 \\
         \hline
         \hline
    \end{tabular}
    \label{tab:diversity}
    (More diverse if the measure is greater ($\uparrow$) or lower ($\downarrow$))
\end{table}

\subsubsection{Effect of selected bags}

Evolutionary algorithms utilise the selection step which is applied in each iteration to produce new offspring by probabilistically selecting individuals from the current generation. One of our goals is to maintain diversity in the data assigned to the bags of the ensemble rather than optimisation.

We implement rank selection in which the bags with the highest fitness are selected as part of the next generation. We conduct the experiment using the Pima dataset, where the number of bags selected during each iteration (generation) is varied. In Table \ref{tab:selection}, we find that the selection step leads to a reduction in accuracy as more bags are selected.

\begin{table}
    \caption{Effect of separate selection step on EvoBagging}
    \centering
    \begin{tabular}{|c|c|c|c|c|c|}
         \hline
         \hline
         Num. Selected Bags & 0 & 4 & 8 & 12 & 16 \\
         \hline
         \hline
         Accuracy & 77.04 & 76.22 & 75.87 & 75.58 & 75.63\\
         \hline
         \hline
    \end{tabular}
    \label{tab:selection}
\end{table}

\begin{figure}[htbp!]
\centering
\includegraphics[scale=0.56]{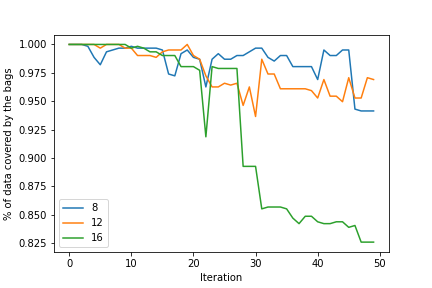}
\caption{The ratio between the number of unique training samples in all bags and the total number of training samples reduces more significantly when the number of selected bags increases from 8 to 16.}
\label{fig:cover}
\end{figure}

Figure \ref{fig:cover} presents the proportion of training data covered \textcolor{black}{by all bags} at each iteration with a different number of selected bags. Clearly, a higher selection rate leads to a constant decrease in data coverage. Over time, such an evolutionary process will only cover a small portion of the data. The reason for this phenomenon is that if a bag has high fitness, it will be more likely to be selected in both the selection step and as a parent for the crossover step. As a result, the data points in this bag will appear more frequently in later generation and eventually dominate the population. An explanation of this is that as fewer data points are covered, the bags will be more similar to each other, which will reduce the overall diversity as discussed in the previous section. In addition, a bag that has high fitness usually contains data points that are easier to learn. Therefore, the final population will only cover the "easy to learn" part of the data which cannot be a good representation of the original training data. % In the scenario with abundant data that can be easily learnt.
However, that is rarely the case in real world applications where labeled data is both scarce and hard to find. An example is the class imbalanced dataset, where applying an evolutionary algorithm with selection step will likely ignore the minority class and optimize only for the majority class. Hence, a more sophisticated fitness function is required to maintain both the classification metric and the balance in learning different classes. 

Intuitively, additional bags in the selection step performs poorly since it is used together with the crossover operator, and adding new or synthetic data instances (i.e. via data augmentation or increasing the rate for generation gap or mutation) can fix the shrinking problem. The selection step tries to retain content (data) of the bag based on their fitness. As bootstrapping allows duplication of a data instances in one bag, the search space of all available bags for this optimization problem will be $N^N$ if there are $N$ data instances, and we limit the maximum bag size to be $N$. By simply keeping/removing the whole bag, we miss the opportunity to optimize at a deeper level by changing the content of the bag and will have to rely on random search of a very large search space. By swapping the data points between two bags based on the fitness, the crossover operator is capable of directly reducing the bias of the two individual learners. %On the other hand, the selection step creates a problem for maintaining diversity by selecting stronger individuals with higher fitness. 
We need diversity in the bags for better performance of the ensemble, rather than providing the bags with similar content. 

\subsubsection{Results on imbalanced datasets}

Apart from existing works where the fitness function is designed specifically for imbalanced datasets \cite{6793456, evobaggingimbalanced, ROSHAN2020103319}, EvoBagging simply optimizes a relatively simple objective defined by the classification performance. Nevertheless, EvoBagging can still perform well on imbalanced datasets as can be seen in Table \ref{tab:result}. In this experiment, we further test its performance on different rates of imbalance between the positive and negative classes for the task of binary classification. To be specific, we perform under-sampling without replacement for the negative class of the Red wine dataset (i.e. the majority class) where 100\%, 75\%, 50\% or 25\% of data points in the majority class remain. On the other hand, all data points in the positive class (i.e. the minority class) are used. \textcolor{black}{Unlike previous experiments, we use the F1 score, which includes precision and recall and is standard for imbalanced datasets, as the classification metric for more informative results.}

\begin{table*}[htbp!]
\caption{Performance of bagging, random forest and EvoBagging for class imbalanced datasets.}
\small
\centering
 \begin{tabular}{| l | l | l | c | c | c | c | c | c | c | c | c |} 
 \hline
 \hline
 Dataset & Imbalance ratio   & Model         & Split                                                             & F1            & Precision     & Recall        & AUC\\
 \hline
 \hline
 {\multirow{18}{*}{Red wine}} & {\multirow{6}{*}{1:1}}             & Bagging       & {\multirow{3}{*}{Train}}   & 100.00 (0)    & 100.00 (0)    & 100.00 (0)    & 100.00 (0)\\ 
  
          &                 & Random forest &                                                                   & 100.00 (0)    & 100.00 (0)    & 100.00 (0)    & 100.00 (0)\\  

          &                 & EvoBagging    &                                                                   & 99.96 (0.01)  & 99.94 (0.01)  & 99.98 (0.01)  & 99.96 (0.01)\\  
 \cline{3-8}
          &                 & Bagging       & {\multirow{3}{*}{Test}}                                           & 83.09 (0.02)  & 97.62 (0.01)  & 72.36 (0.03)  & 82.66 (0.01)\\ 
  
          &                 & Random forest &                                                                   & 79.12 (0.02)  & \textbf{98.48 (0.01)}  & 66.15 (0.02)  & 81.04 (0.01)\\  

          &                 & EvoBagging    &                                                                   & \textbf{85.28 (0.02)}  & 97.49 (0.01)  & \textbf{75.86 (0.04)}  & \textbf{84.03 (0.02)}\\  
 \cline{2-8}
          & {\multirow{6}{*}{1:3.68}}          & Bagging       & {\multirow{3}{*}{Train}}                       & 99.97 (0.01)  & 100.00 (0)    & 99.95 (0.01)  & 99.98 (0.01)\\ 
  
          &                 & Random forest &                                                                   & 99.98 (0.01)  & 100.00 (0)    & 99.97 (0.01)  & 99.98 (0.01)\\  

          &                 & EvoBagging    &                                                                   & 99.77 (0.01)  & 99.97 (0.01)  & 99.57 (0.01)  & 99.78 (0.01)\\ 
 \cline{3-8}          
          &                 & Bagging       & {\multirow{3}{*}{Test}}                                           & 74.10 (0.04)  & 94.09 (0.01)  & 61.27 (0.05)  & 78.56 (0.03)\\ 
  
          &                 & Random forest &                                                                   & 70.64 (0.03)  & 95.45 (0.01)  & 56.13 (0.04)  & 76.65 (0.02)\\  

          &                 & EvoBagging    &                                                                   & \textbf{78.55 (0.01)}  & \textbf{95.53 (0.01)}  & \textbf{66.77 (0.02)}  & \textbf{81.65 (0.02)}\\ 
 \cline{2-8}
          & {\multirow{6}{*}{1:6.35}}          & Bagging       & {\multirow{3}{*}{Train}}                       & 99.83 (0.01)  & 100.00 (0)    & 99.66 (0.01)  & 99.83 (0.01)\\ 
  
          &                 & Random forest &                                                                   & 99.88 (0.01)  & 100.00 (0)    & 99.77 (0.01)  & 99.89 (0.01)\\  

          &                 & EvoBagging    &                                                                   & 99.53 (0.01)  & 99.94 (0.01)  & 99.13 (0.01)  & 99.56 (0.01)\\
 \cline{3-8}            
          &                 & Bagging       & {\multirow{3}{*}{Test}}                                           & 67.41 (0.02)  & 75.52 (0.02)  & 63.02 (0.02)  & 79.65 (0.01)\\ 
  
          &                 & Random forest &                                                                   & 66.35 (0.03)  & 75.73 (0.03)  & 58.60 (0.03)  & 77.90 (0.02)\\  

          &                 & EvoBagging    &                                                                   & \textbf{69.99 (0.01)}  & \textbf{76.57 (0.02)}  & \textbf{65.11 (0.02)}  & \textbf{80.93 (0.01)}\\
 \hline
 \hline
 {\multirow{12}{*}{Pima}}     & {\multirow{6}{*}{1:1}}             & Bagging       & {\multirow{3}{*}{Train}}   & 99.99 (0.01)  & 99.99 (0.01)  & 99.99 (0.01)  & 99.99 (0.01)\\ 
  
          &                 & Random forest &                                                                   & 100.00 (0)    & 100.00 (0)    & 100.00 (0)    & 100.00 (0)\\  

          &                 & EvoBagging    &                                                                   & 99.87 (0.01)  & 99.90 (0.01)  & 99.85 (0.01)  & 99.88 (0.01)\\  
 \cline{3-8}            
          &                 & Bagging       & {\multirow{3}{*}{Test}}                                           & 71.27 (0.03)  & 93.13 (0.01)  & 57.84 (0.04)  & 70.42 (0.02)\\ 
  
          &                 & Random forest &                                                                   & 67.76 (0.04)  & 93.14 (0.01)  & 53.39 (0.05)  & 68.87 (0.02)\\  

          &                 & EvoBagging    &                                                                   & \textbf{71.70 (0.01)}  & \textbf{93.81 (0.01)}  & \textbf{58.02 (0.01)}  & \textbf{71.36 (0.01)}\\  
 \cline{2-8}
          & {\multirow{6}{*}{1:3.68}}          & Bagging       & {\multirow{3}{*}{Train}}                       & 99.93 (0.01)  & 99.98 (0.01)  & 99.88 (0.01)  & 99.94 (0.01)\\ 
  
          &                 & Random forest &                                                                   & 99.96 (0.01)  & 100.00 (0)    & 99.93 (0.01)  & 99.96 (0.01)\\  

          &                 & EvoBagging    &                                                                   & 99.25 (0.01)  & 99.67 (0.01)  & 98.83 (0.01)  & 99.33 (0.01)\\  
 \cline{3-8}            
          &                 & Bagging       & {\multirow{3}{*}{Test}}                                           & 61.40 (0.03)  & 66.94 (0.03)  & 56.76 (0.02)  & 70.78 (0.01)\\ 
  
          &                 & Random forest &                                                                   & 58.92 (0.03)  & 68.64 (0.03)  & 51.67 (0.03)  & 69.46 (0.02)\\  

          &                 & EvoBagging    &                                                                   & \textbf{64.52 (0.01)}  & \textbf{70.36 (0.01)}  & \textbf{59.63 (0.02)}  & \textbf{73.01 (0.01)}\\  
 \hline
 \hline
 {\multirow{22}{*}{Abalone}}  & {\multirow{6}{*}{1:1}}             & Bagging       & {\multirow{3}{*}{Train}} & 100.00 (0)    & 100.00 (0)    & 100.00 (0)    & 100.00 (0)\\ 
  
          &                 & Random forest &                                                                   & 100.00 (0)    & 100.00 (0)    & 100.00 (0)    & 100.00 (0)\\ 

          &                 & EvoBagging    &                                                                   & 99.96 (0.01)  & 99.93 (0.01)  & 100.00 (0.01) & 99.96 (0.01)\\ 
 \cline{3-8}          
          &                 & Bagging       & {\multirow{3}{*}{Test}}                                           & 39.35 (0.04)  & 97.35 (0.01)  & 24.73 (0.03)  & 61.04 (0.01)\\ 
  
          &                 & Random forest &                                                                   & 31.29 (0.03)  & 97.40 (0.01)  & 18.68 (0.02)  & 58.36 (0.01)\\  

          &                 & EvoBagging    &                                                                   & \textbf{50.26 (0.02)}  & \textbf{97.78 (0.02)}  & \textbf{33.84 (0.02)}  & \textbf{65.40 (0.01)}\\ 
 \cline{2-8}
          & {\multirow{6}{*}{1:5.66}}          & Bagging       & {\multirow{3}{*}{Train}}                       & 99.95 (0.01)  & 99.99 (0.01)  & 99.91 (0.01)  & 99.95 (0.01)\\ 
  
          &                 & Random forest &                                                                   & 99.99 (0.01)  & 100.00 (0)    & 99.98 (0.01)  & 99.99 (0.01)\\  

          &                 & EvoBagging    &                                                                   & 99.80 (0.01)  & 99.93 (0.01)  & 99.67 (0.01)  & 99.83 (0.01)\\  
 \cline{3-8}          
          &                 & Bagging       & {\multirow{3}{*}{Test}}                                           & 34.69 (0.04)  & 86.63 (0.02)  & 21.75 (0.03)  & 59.71 (0.01)\\ 
  
          &                 & Random forest &                                                                   & 23.23 (0.05)  & 85.78 (0.04)  & 13.51 (0.01)  & 56.00 (0.02)\\  

          &                 & EvoBagging    &                                                                   & \textbf{41.29 (0.02)}  & \textbf{91.87 (0.01)}  & \textbf{26.64 (0.01)}  & \textbf{64.49 (0.01)}\\  
 \cline{2-8}
          & {\multirow{6}{*}{1:10.32}}         & Bagging       & {\multirow{3}{*}{Train}}                       & 99.87 (0.01)  & 99.98 (0.01)  & 99.75 (0.01)  & 99.88 (0.01)\\ 
  
          &                 & Random forest &                                                                   & 99.93 (0.01)  & 100.00 (0)    & 99.85 (0.01)  & 99.93 (0.01)\\  

          &                 & EvoBagging    &                                                                   & 97.64 (0.01)  & 99.83 (0.01)  & 95.58 (0.01)  & 97.78 (0.01)\\  
 \cline{3-8}          
          &                 & Bagging       & {\multirow{3}{*}{Test}}                                           & 30.81 (0.06)  & 79.97 (0.05)  & 19.21 (0.04)  & 58.71 (0.02)\\ 
  
          &                 & Random forest &                                                                   & 15.28 (0.04)  & 75.06 (0.06)  & 8.56 (0.03)   & 53.76 (0.01)\\  

          &                 & EvoBagging    &                                                                   & \textbf{34.60 (0.01)}  & \textbf{84.37 (0.01)}  & \textbf{21.78 (0.02)}  & \textbf{60.11 (0.01)}\\  
 \cline{2-8}
          & {\multirow{6}{*}{1:15}}            & Bagging       & {\multirow{3}{*}{Train}}                       & 99.54 (0.01)  & 100.00 (0)    & 99.09 (0.01)  & 99.55 (0.01)\\ 
  
          &                 & Random forest &                                                                   & 99.65 (0.01)  & 100.00 (0)    & 99.31 (0.01)  & 99.65 (0.01)\\  

          &                 & EvoBagging    &                                                                   & 88.90 (0.06)  & 99.67 (0.01)  & 80.67 (0.09)  & 90.33 (0.05)\\  
 \cline{3-8}          
          &                 & Bagging       & {\multirow{3}{*}{Test}}                                           & 27.06 (0.02)  & 41.57 (0.02)  & 20.10 (0.02)  & 59.11 (0.01)\\ 
  
          &                 & Random forest &                                                                   & 16.76 (0.04)  & 38.21 (0.08)  & 10.77 (0.03)  & 54.81 (0.02)\\  

          &                 & EvoBagging    &                                                                   & \textbf{28.60 (0.02)}  & \textbf{48.16 (0.04)}  & \textbf{20.38 (0.02)}  & \textbf{59.46 (0.01)}\\  

  \hline   
   \hline
\end{tabular}
\label{tab:imbalanced}
\end{table*}

\textcolor{black}{Table \ref{tab:imbalanced} shows that EvoBagging dominates the baseline models on  the given class  imbalance ratio settings. An important observation is that with higher imbalance ratio, there is significant  improvement of EvoBagging when  compared to the baseline methods (bagging and random forests).} There is clear evidence of the effectiveness of EvoBagging in strongly imbalanced datasets although no mechanisms in the algorithm is specifically designed for this purpose. The reason for this effect is the ability of the algorithm to optimize towards a representative set of bags using its operators, namely the crossover, mutation, and the selection with generation gap. These operators directly update the content of a bag for a higher fitness value. As the classification metric (F1 in this case), plays a major role in the fitness function; the second term $\frac{k +  \phi_b}{k}$ remains rather stable after a few iterations as a result of stable bag sizes. We find that the classification for both majority and minority classes are improved.

\subsubsection{Effect of the voting rule}

The voting rule plays a crucial role in the task of classification using an ensemble machine learning model as the predictions of individual learners must be aggregated for the final output. Hence, majority voting is commonly implemented while weighted voting based on the individual learner's performance is slightly more complicated with more computation required. To build an ensemble machine learning model, one usually has to determine which voting rule to use in order to limit  costly hyperparameter tuning. In situations where there are computational constraints and stable results are required, it is especially beneficial for a machine learning model not to be highly prone to changes in the hyper-parameters, including the choice of voting rule in this situation.

\begin{figure}[!h]
    \centering
    \begin{subfigure}[t]{.45\textwidth}
        \centering
        \includegraphics[width=\textwidth]{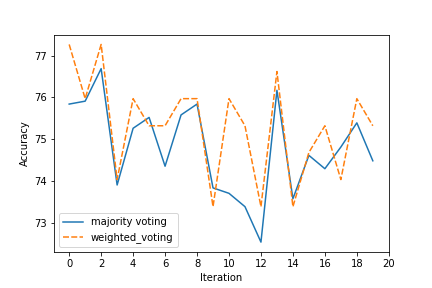}
        \caption{Pima}
        \label{fig:voting_pima} 
    \end{subfigure}
    \begin{subfigure}[t]{.45\textwidth}
       \centering
       \includegraphics[width=\textwidth]{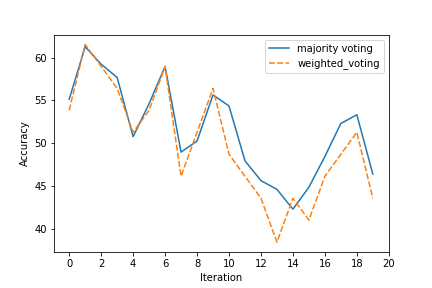}
       \caption{Two-spiral}
       \label{fig:voting_spiral}
    \end{subfigure}
    \caption{Compare classification accuracy between two different voting rules for each iteration}
    \label{fig:voting}
\end{figure}

Fortunately, our experiments with EvoBagging have shown that its performance is not strongly affected by the voting rule used. On Pima and Two-spiral datasets, we record the accuracy of the aggregated prediction on the test set using both majority voting and weighted voting for each iteration. In weighted voting, the aggregated prediction for a data point is determined by the following formula:

\begin{equation}
    \hat{y} = \begin{cases}
        0, & \text{if }  \sum_{l \in L} P_{neg}acc_l < \sum_{l \in L} P_{pos}acc_l \\
        1, & \text{other wise}
    \end{cases}
\end{equation}

where $l$ is an individual learner in the set $L$, and $acc_l$ is the training accuracy of $l$.

In Figure \ref{fig:voting_pima}, we observe that the accuracy scores of both majority voting and weighted voting are mostly similar over time (iterations) for the Pima dataset (slightly class imbalanced). When the classes are perfectly balanced in the two-spiral problem, the result is unchanged (see Figure \ref{fig:voting_spiral}). The voting rule has little to no impact on the performance of the algorithm; hence, majority voting can be used for faster inference and no tuning is necessary for the choice of voting rule.

\section{Discussion}

The most important part of the EvoBagging algorithm is the crossover operation. By constantly replacing incorrectly classified samples with new samples, it is expected that a bag will eventually contain samples that can be effectively modeled. Note that these samples are not essentially adjacent in a representation space. They can be seen as those that follow a simpler set of rules than the random bootstrapped samples. There is considerable support in the literature for the importance of crossover in obtaining convergence \cite{jansen2002analysis,doerr2012crossover}. It has been shown in the case of genetic algorithms that crossover plays a role in diversification \cite{qi1994theoretical}. It has also been shown that crossover can improve convergence with bagging in the presence of imbalanced datasets \cite{ROSHAN2020103319}. The computational results are therefore consistent and support theoretical literature in particular \cite{ROSHAN2020103319}. 

We note that the proposed framework provides reasonably accurate results using single objective optimization when compared to other methods from the literature. In future works, given more complicated problems, we can approach the proposed framework using a multi-objective optimization approach since they have shown to be promising in ensemble learning. \textcolor{black}{EvoBagging considers the optimisation of the ensemble of bags using an evolutionary algorithm where the population of individuals represents the  data in the bags which are evolved by shuffling the indices. Hence, this is a discrete parameter optimisation problem that improved the content of the bags for better individual learning models (bags).} Hence, EvoBagging has the ability to  optimize the grouping of the data featured in the bags, unlike bagging which considers the bag to be fixed in the ensemble. EvoBagging can also be used as a post-processing step to gain better results from bagging or random forest. The design of EvoBagging does not require heavy hyper-parameter tuning, and a global convergence is guaranteed due to global search properties of evolutionary algorithms. 

EvoBagging has only been evaluated on classification tasks and its effectiveness on regression tasks can be explored in future work. Furthermore, in our implementation, decision trees are used as individual learners for the ensemble in EvoBagging. The effectiveness of EvoBagging can be further verified for other types of learners such as neural networks; however, this could be computationally costly. 

The computational requirement of EvoBagging is much larger than that of bagging or random forest. Let $T$ be the amount of time to train an individual learner such as a decision tree or an artificial neural network. The time complexity of bagging and random forest will be $O(N\times T)$ since bootstrapping is generally faster than training a machine learning model. $O(N\times T)$ is also the time complexity of each iteration of EvoBagging where each individual learner will be trained on a bag. The reason is that the time complexities of operators like generation gap (multiple bootstrappings), crossover (swapping data indices between bags), and mutation (swapping data indices between a bag and the remaining data) are again minimal compared to training individual learners. Therefore, if an EvoBagging algorithm takes $I$ iterations to converge, its time complexity will be $O(I\times N\times T)$. To be scalable for larger datasets, future works should focus on methods to reduce the computational requirement of EvoBagging by either optimising with fewer iterations or approximating the fitness score without a full evaluation of all the bags.

\section{Conclusions}

In this paper, we presented a novel implementation of evolutionary algorithm for ensemble learning known as EvoBagging. We provided an extensive evaluation on multiple datasets to evaluate the effectiveness and to provide rationale for the algorithm design. Our results show that EvoBagging improves the bias (error) in the individual learners from the bags that are sampled with bootstrapping.

The results have shown that EvoBagging successfully outperforms bagging and random forests for both binary and multi-class classification problems. The results have also shown that EvoBagging can maintain good performance on both class balanced and imbalanced datasets. The design choices for EvoBagging components, such as crossover, mutation, and generation gap-based selection have demonstrated their relevance by providing improved performance accuracy.  EvoBagging maintains a diverse ensemble of individual learners which is a major factor for the improved performance when compared to conventional ensemble methods such as bagging and random forests. %We also demonstrated that removing the separate selection step, which is a major difference compared to the canonical evolutionary algorithm, has brought a significant improvement by providing diversity in the individuals. 

\section*{Code and data}

The code and all datasets for reproducing the experiments are available at: https://github.com/sydney-machine-learning/evolutionary-bagging.

\bibliographystyle{elsarticle-num-names} 
\bibliography{cas-refs}





% \end{thebibliography}
\end{document}